\documentstyle[AAAI]{article}
\input{psfig}

\newcommand{\datac}{\mathrm{DC}}

\newcommand{\at}{{\mathit At}}

\newtheorem{theorem}{Theorem}

\begin{document}
\title{DATALOG with constraints --- an answer-set programming system}
\author{Deborah East \and Miros\l aw Truszczy\'nski\\
		Department of Computer Science\\
		University of Kentucky\\
		Lexington KY 40506-0046, USA \\
		email: deast$|$mirek@cs.uky.edu}

\maketitle

\begin{abstract}
\begin{quote}
Answer-set programming (ASP) has emerged recently as a viable programming
paradigm well attuned to search problems in AI, constraint satisfaction
and combinatorics. Propositional logic is, arguably, the simplest ASP 
system with an intuitive
semantics supporting direct modeling of problem constraints. However, for 
some applications, especially those requiring that transitive closure 
be computed, it requires additional variables and results in large theories. 
Consequently, it may not be a practical computational tool for such problems.
On the other hand, ASP systems based on nonmonotonic logics, such as
stable logic programming, can handle
transitive closure computation efficiently and, in general, yield
very concise theories as problem representations. Their semantics is,
however, more complex. Searching for the middle ground, in this paper 
we introduce a new nonmonotonic logic, {\em DATALOG with constraints} 
or $\datac$. Informally, $\datac$ theories consist of propositional clauses 
(constraints) and of Horn rules. The 
semantics is a simple and natural extension of the semantics 
of the propositional logic. However, thanks to the presence of Horn
rules 
in the system, modeling of transitive closure becomes straightforward.
We describe the syntax and semantics of $\datac$, and study its
properties. We discuss an implementation of $\datac$ and present results 
of experimental study of the effectiveness of $\datac$, comparing it with 
the {\tt csat} satisfiability checker and {\tt smodels} implementation of 
stable logic programming. Our results show that $\datac$ is competitive
with the other two approaches, in case of many search problems, often
yielding much more efficient solutions.\\
{\bf Content Areas:} constraint saitsfaction, search, knowledge representation,
logic programming, nonmonotonic reasoning.
\end{quote}
\end{abstract}

\section{Introduction}\label{intro}

Many important computational problems in combinatorial optimization,
constraint satisfaction and artificial intelligence can be cast as 
search problems. Answer-set programming (ASP) \cite{mt99,nie98} was 
recently identified as a declarative programming paradigm appropriate
for such applications. Logic programming with the stable-model
semantics ({\em stable logic programming}, for short)
was proposed as an embodiment of this paradigm. 
Disjunctive logic programming with 
the answer-set semantics is another implementation of ASP currently 
under development \cite{elmps98}. Early experimental results demonstrate 
the potential of answer-set programming approaches in such areas as 
planning and constraint satisfaction \cite{nie98,li99a,li99b}.

In this paper we describe another formalism that implements the ASP 
approach. We call it {\em DATALOG with constraints} and denote by 
$\datac$. Our goal is to design an ASP system with a semantics more 
readily understandable than the semantics of stable models. We seek 
a semantics that would be as close as possible to
propositional satisfiability yet as expressive and as 
effective, especially from the point of view of conciseness of 
representations and time performance, as the stable logic programming.
We argue that $\datac$ has a potential to become a practical
declarative programming tool. We show that it yields intuitive
and small-size encodings, we characterize its complexity and expressive 
power and present computational experiments demonstrating its effectiveness.  

Answer-set programming is a paradigm in which programs are built as
theories in some formal system $\cal F$ with a well-defined syntax, and 
with a semantics that assigns to a theory $P$ in the system a 
{\em collection} of subsets of some domain.
These subsets are referred to as {\em answer sets}
of $P$ and specify the results of computation based on $P$. To
solve a problem $\Pi$ in an ASP formalism, we find a program
$P$ so that the solutions to $\Pi$ can be reconstructed, in polynomial
(ideally, linear) time, from the answer sets to $P$. 

The definition of the answer-set programming given above is very
general. Essentially any logic formalism can be a basis for an answer-set
programming system. For instance, the propositional logic gives
rise to an ASP system: programs are collections of propositional
clauses, their models are answer sets. To solve, say, a planning problem,
we encode the constraints of the problem as propositional clauses
in such a way that legal plans are determined by models of the resulting
propositional theory. This approach, called {\em satisfiability
planning}, received significant attention lately and was shown to be 
quite effective \cite{sk92,ks96,kms96}.

Recently, several implementations of the ASP approach were developed
that are based on nonmonotonic logics such as {\tt smodels} \cite{ns96}, 
for stable logic programming, {\tt dlv} \cite{elmps98}, for disjunctive 
logic programming with answer-set semantics, and 
{\tt deres} \cite{cmt96}, for default logic with Reiter's extensions.
All these systems have been extensively studied. Promising 
experimental results concerning their performance were reported
\cite{cmmt99,elmps98,nie98}. 
 
The question arises which formal logics are appropriate as bases
of answer-set programming implementations. To discuss such a general 
question one needs to formulate quality criteria with respect to which 
ASP systems can be compared. At the very least, these criteria should
include:\\
1. expressive power\\
2. time performance\\
3. simplicity of the semantics\\
4. ease of coding, conciseness of programs.

We will discuss these criteria in detail elsewhere. We will make 
here only a few brief comments on the matter. From the point of view of 
the expressive power all the systems that we discussed are quite 
similar. Propositional logic and stable logic programming are 
well-attuned to the class NP \cite{sch95}. Disjunctive logic 
programming and default logic capture the class $\Sigma_P^2$ 
\cite{eg95,ceg97}. However, this distinction is not essential as 
recently pointed out in \cite{jnsy00}. The issue of time performance 
can be resolved only through comprehensive experimentation and this work 
is currently under way. 

As concerns inherent complexity of the system and intuitiveness of the
semantics, ASP systems based on the propositional logic seem to be clear 
winners. However, propositional logic is monotone and modeling indefinite 
information and phenomena such as the frame problem is not quite 
straightforward. In applications involving the
computation of transitive closures, as in the problem of existence of
hamilton cycles, it leads to programs that are large and, thus, difficult 
to process. In this respect, ASP systems based on nonmonotonic logics 
have an edge. They were designed to handle incomplete and indefinite 
information. Thus, they often yield more concise programs. However, 
they require more elaborate formal machinery and their semantics are more 
complex.

Searching for the middle ground between systems such as logic programming
with stable model semantics and propositional logic, we propose here a
new ASP formalism, $\datac$. Our guiding principle was to design a
system which would lead to small-size encodings, believing that small
theories will lead to more efficient solutions. We show that $\datac$ is 
nonmonotonic, has the same expressive power as stable logic programming 
but that its semantics stays closer to that of propositional logic. Thus, 
it is arguably simpler than the stable-model semantics. We present 
experimental results that demonstrate that $\datac$ is competitive with 
ASP implementations based on nonmonotonic logics (we use {\tt smodels} 
for comparison) and those based on propositional logics (we use 
{\tt csat} \cite{dub96} in our experiments). Our results strongly indicate
that formalisms which provide smaller-size encodings  are more
effective as practical search-problem solvers.

\section{DATALOG with constraints}

A $\datac$ theory (or program) consists of constraints and Horn rules
(DATALOG program). This fact motivates out choice of terminology ---
DATALOG with constraints. We start a discussion of $\datac$ with the 
propositional case. Our language 
is determined by a set of atoms $\at$. We will assume that $\at$ is of the 
form $\at = \at_C \cup \at_H$, where $\at_C$ and $\at_H$ are disjoint. 

A {\em DC theory} (or {\em DC program}) is a triple $T=(T_C,T_H,T_{PC})$, where\\ 
1. $T_C$ is a set of propositional clauses $\neg a_1\vee\ldots\vee \neg a_m
\vee b_1\vee\ldots\vee b_n$ such that all $a_i$ and $b_j$ are from 
$\at_C$,\\ 
2. $T_H$ is a set of Horn rules $a_1\wedge\ldots\wedge a_m\rightarrow b$
such that $b\in \at_H$ and all $a_i$ are from $\at$,\\
3. $T_{PC}$ is a set of clauses over $\at$.\\
By $\at(T)$, $\at_C(T)$ and $\at_{PC}(T)$ we denote the set of atoms from
$\at$, $\at_C$ and $\at_{PC}$, respectively, that actually appear in $T$.

With a $\datac$ theory $T = (T_C,T_H,T_{PC})$ we associate a family of 
subsets of $\at_C(T)$. We say that a set $M\subseteq \at_C(T)$ {\em satisfies} 
$T$ (is an {\em answer set} of $T$) if\\ 
1. $M$ satisfies all the clauses in $T_C$, and\\
2. the closure of $M$ under the Horn rules in $T_H$,
$M^c=LM(T_H\cup M)$ satisfies all clauses in $T_{PC}$ ($LM(P)$ denotes
the least model of a Horn program $P$).\\
Intuitively, the collection of clauses in $T_C$ can 
be thought of as a representation of the constraints of the problem, 
Horn rules in $T_H$ can be viewed as a mechanism to compute closures of sets 
of atoms satisfying the constraints in $T_C$, and the clauses in $T_{PC}$ can 
be regarded as constraints on closed sets (we refer to them as
{\em post-constraints}). A set of atoms $M \subseteq \at_C(T)$ is a model 
if it (propositionally) satisfies the constraints in $T_C$ and if its 
closure (propositionally) satisfies the constraints in $T_{PC}$. Thus, 
the semantics of $\datac$ retains much of the simplicity of the semantics 
of propositional logic.   

$\datac$ can be used as a computational tool to solve search problems. 
We define a search problem $\Pi$ to be determined
by a set of finite {\em instances}, $D_\Pi$, such that for each instance 
$I \in D_\Pi$, there is a finite set $S_\Pi(I)$ of all
{\em solutions} to $\Pi$ for the instance $I$. For example, the problem
of finding a hamilton cycle in a graph is a search problem: graphs are
instances and for each graph, its hamilton cycles (sets of their edges)
are solutions. A $\datac$ theory $T = (T_C,T_H,T_{PC})$ {\em solves}
a search problem $\Pi$ if solutions to $\Pi$ can be computed (in
polynomial time) from answer sets to $T$. Propositional logic and stable
logic programming are used as problem solving formalisms following the
same general paradigm. To illustrate all the concepts introduced here and 
show how $\datac$ programs can be built by modeling
problem constraints, we will now present a $\datac$ program that solves the
hamilton-cycle problem.

Consider a directed graph $G$ with the vertex set $V$ and the edge set $E$. 
Consider a set of atoms $\{hc(a,b)\colon (a,b)\in E\}$. An intuitive
interpretation of an atom $hc(a,b)$ is that the edge $(a,b)$ is in a 
hamilton cycle. Include in $T_C$ all clauses of the form 
$\neg hc(b,a)\vee \neg hc(c,a)$, where $a,b,c\in V$, $b\not=c$ and 
$(b,a), (c,a) \in E$. In addition,
include in $T_C$ all clauses of the form $\neg hc(a,b)\vee \neg
hc(a,c)$, where $a,b,c\in V$, $b\not=c$ and $(a,b), (a,c) \in E$. 
Clearly, the set of propositional variables of the form $\{hc(a,b)\colon
(a,b)\in F\}$, where $F\subseteq E$, satisfies all clauses in $T_C$ if 
and only if no two 
distinct edges in $F$ end in the same vertex and no two distinct edges 
in $F$ start in the same vertex. In other words, $F$ spans a collection 
of paths and cycles in $G$. 

To guarantee that the edges in $F$ define a hamilton cycle, we must
enforce that all vertices of $G$ are reached by means of the edges in $F$
if we start in some (arbitrarily chosen) vertex of $G$. This can be
accomplished by means of a simple Horn program. Let us choose a vertex,
say $s$, in $G$. Include in $T_H$ the Horn rules $hc(s,t)\rightarrow
vstd(t)$, for every edge $(s,t)$ in $G$. In addition, include in 
$T_H$ Horn rules $vstd(t),hc(t,u)\rightarrow vstd(u)$, for every edge
$(t,u)$ of $G$ not starting in $s$. Clearly, the least model of $F\cup
T_H$, where $F$ is a subset of $E$, contains precisely these 
variables of the form $vstd(t)$ for which $t$ is reachable from $s$ by a
{\em nonempty} path spanned by the edges in $F$. Thus, $F$ is the set of
edges of a hamilton cycle of $G$ if and only if the least model of
$F\cup T_H$, contains variable $vstd(t)$ for every vertex $t$ of $G$.  
Let us define $T_{PC} =\{vstd(t)\colon t\in V\}$ and
$T_{ham}(G)=(T_C,T_H,T_{PC})$. It follows that hamilton cycles of $G$ 
can be reconstructed (in linear time) from answer sets to the $\datac$ 
theory $T_{ham}(G)$. In other words, to find a hamilton cycle in $G$, it
is enough to find an answer set for $T_{ham}(G)$.

This example illustrates the simplicity of the semantics --- it is only
a slight adaptation of the semantics of propositional logic to the case
when in addition to propositional clauses we also have Horn rules in 
theories. It also illustrates the power of $\datac$ to generate concise
encodings. All known propositional encodings of the hamilton-cycle
problem require that additional variables are introduced to ``count''
how far from the starting vertex an edge is located. Consequently, 
propositional encodings are much larger and lead to inefficient 
computational approaches to the problem. We present experimental 
evidence to this claim later in the paper.

The question arises which search problems can be represented 
(and solved) by means of finding answer sets to appropriate $\datac$ 
programs. In general, the question remains open. We have an answer,
though, if we restrict our attention to the special case of decision 
problems.
Consider a $\datac$ theory $T = (T_C,T_H,T_{PC})$, where $T_H=T_{PC}=
\emptyset$. Clearly, $M$ is an answer set for $T$ if and only if $M$ is
a model of the collection of clauses $T_C$. Thus, the problem of
existence of an answer set is at least as hard as the propositional
satisfiability problem. On the other hand, for every $\datac$ theory $T$
and for every set $M\subseteq \at_C(T)$, it can be checked in linear
time whether $M$ is an answer set for $T$. Thus, we obtain the
following complexity result.

\begin{theorem}
The problem of existence of an answer set for a finite propositional
$\datac$ theory $T$ is NP-complete.  
\end{theorem}

It follows that every problem in NP can be polynomially reduced to
the problem of existence of an answer set for a propositional $\datac$
program. Thus, given a problem $\Pi$ in NP, for every instance $I$
of $\Pi$, $\Pi$ can be decided by deciding the existence of an answer 
set for the $\datac$ program corresponding to $\Pi$ and $I$.

Propositional $\datac$ can be extended to the predicate case. It is
important as it significantly simplifies the task of developing programs
for solving problems with $\datac$. In the example discussed above, 
the theory $T_{ham}(G)$ depends heavily on the input. Each time we
change the input graph, a different theory has to be used. However, 
when constructing predicate $\datac$-based solutions to 
a problem $\Pi$, it is often possible to separate the representation of 
an instance (input) to $\Pi$ from that of the constraints that define 
$\Pi$. As a result only one (predicate) program describing the constraints 
of $\Pi$ needs to be written. Specific input for the program, say
$I$, can be described separately as a collection of facts (according to
some uniform schema). Both parts together can be combined to yield a 
$\datac$ program whose answer sets determine solutions to $\Pi$ for 
the input $I$. Such an approach, we will refer to it as {\em uniform},
is often used in the context of 
DATALOG, DATALOG$^\neg$ or logic programming to study complexity of
these systems as query languages. The part representing input is
referred to as the {\em extensional} database. The part representing
the query or the problem is called the {\em intensional} database or
program. Due to the space limitations we do not discuss the details of 
the predicate case here. They will be given in the full version of the 
paper. We only state a generalization of Theorem 1.

\begin{theorem}
The expressive power of $\datac$ is the same as that of stable logic
programming. In particular, a decision problem $\Pi$ can be solved 
uniformly in $\datac$ if and only if $\Pi$ is in the class {\rm NP}.
\end{theorem}

\section{Implementation}

Some types of constraints appear frequently in applications. For instance, 
when defining plans we may want to specify a constraint that says that
exactly one action from the set of allowed actions be selected at each
step. Such constraints can be modeled by {\em collections} of clauses. To 
make sure $\datac$ programs are as easy to write and as concise as possible 
we have extended the syntax of $\datac$ by providing explicit ways to
model constraints of the form ``select at least (at most, exactly) $k$
elements from a set''. Having these constraints results in shorter
programs which, as we believe, has a significant positive effect of the
performance of our system. 

An example of a select constraint with a short explanation is presented here.
Let $PRED$ be the set of predicates occurring in the IDB.
For each variable $X$ declared in the IDB  the range $R(X)$ of $X$
is determined by the EDB.

\begin{description}
    \item[Select$(n,m,\vec{Y};p_1(\vec{X}),\dots,p_i(\vec{X},\vec{Y}))
q(\vec{X},\vec{Y})$,]
        where $n,m$ are nonnegative integers such that $n \leq m, q \in PRED$ and
        $p_1, \dots ,p_i$ are
        EDB predicates or logical conditions (logical conditions can be comparisons of
       arithmetic expressions or string comparisons).
        The interpretation of this constraint is as follows:
        for every $\vec{x} \in R(\vec{X})$
        at least $n$ atoms and at most $m$ atoms in the set
        $\{q(\vec{x},\vec{y}): \vec{y} \in R(\vec{Y}) \}$ are true.
\end{description}

We implemented $\datac$ in the predicate setting. Thus, our system
consists of two main modules. The first of them, referred to as {\tt
grounder}, converts a predicate $\datac$ program (consisting of both 
the extensional and intensional parts) into the corresponding
propositional $\datac$ program. The second module, $\datac$ {\em
solver}, denoted {\tt dcs}, finds the answer sets to propositional 
$\datac$ programs. Since we focus on the propositional case here, 
we only describe the key ideas behind the $\datac$ solver, {\tt dcs}.
 
The $\datac$ solver uses a Davis-Putnam type approach, with
backtracking, propagation and lookahead (also called literal 
testing), to deal with constraints represented as clauses, {\em select} 
constraints and Horn rules, and to search for answer sets. The lookahead 
in $\datac$ is similar to local processing performed in {\tt csat}
\cite{dub96}.
However, we use different methods to determine how many literals to
consider in the lookahead phase. Other techniques, especially 
propagation and search heuristics, were designed specifically for the
case of $\datac$ as they must take into account the presence of Horn 
rules in programs.

The lookahead procedure selects a number of literals which have not 
yet been assigned a value. For each such literal, the procedure tries
both truth values: true and false. For each assignment, the theory is 
evaluated using propagation. If in both cases a contradiction is
reached, then it is necessary to backtrack. If for only one evalution
a conflict is reached, then the literal is assigned the other truth value and
we proceed to the next step. If neither evaluation results in a
contradiction, we cannot assign a truth value to this literal 
but we save the data such as the number of forced literals 
and the number of clauses satisfied, computed during propagation.  

Clearly, if all unassigned literals were tested it would prune
the most search space. At the same time, the savings might not be 
large enough to compensate for the increase in the running time caused 
by extensive lookahead. Thus, we select only a portion of all unassigned
literals for lookahead. The number of literals to consider was
established empirically (it does not depend on the size of the theory). 
Since not all literals are selected, it is important 
to focus on those literals that are likely to result in a contradiction 
for at least one of the truth values. In our implementation, we select the 
most constrained literals, as determined by their weights. 

Specifically, each constraint is assigned a weight based on its current
length and types (recall that in addition to propositional clauses, we
also allow other types of constraints, e.g., select constraints). 
The shorter the constraint the greater its weight. Also, certain types 
of constraints force more assignments on literals and are given a greater 
weight than other 
constraints of the same length. Every time a literal appears in an 
unsatisfied constraint, the weight of that literal is incremented by 
the weight of the clause.  

After testing a predetermined number of literals without finding a forced truth
assignment and without backtracking, the information computed
during propagation is used to choose the 
next literal for which both possible truth assignments have to be tested
(branching literal). The choice of the next branching literal is based 
on an approximation of which literal, once assigned a truth value, will 
force the truth assignments onto the largest number of other literals 
and will satisfy the largest number of clauses. Using the data computed during 
propagation gives more 
accurate information on which to base such approximations. The methods 
used for determining which literals to select in the lookahead and which 
data to collect and save during the propagation phase are two key 
ways in which the literal testing procedure differs from the local 
processing of {\tt csat}.

\section{Experimentation}
\label{exps}

We compared the performance of $\datac$ solver {\tt dcs} with {\tt smodels}, 
a system for computing stable models of logic programs \cite{ns96}, and 
{\tt csat}, a system for testing propositional satisfiability \cite{dub96}. 
In the case of {\tt smodels} we used version 2.24 in conjunction with
the grounder {\tt lparse}, version 0.99.41. These versions
of {\tt lparse} and {\tt smodels} implement the expressive rules described 
by \cite{sim99}. The expressive rules were used whenever applicable during
the testing.
The programs were all executed on a Sun SparcStation 
20. For each test we report the cpu user times for processing the
corresponding propositional program or theory. We tested all three system
to compute hamilton cycles and colorings in graphs, to solve the
$N$-queens problem, to prove that the pigeonhole problem has no solution
if the number of pigeons exceeds the number of holes, and to compute
Schur numbers. 

\begin{figure}
\centerline{\hbox{\psfig{figure=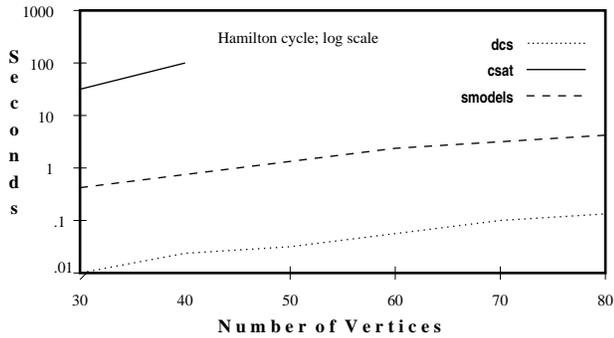}}}
\caption{Hamilton cycle problem; times on the log scale as
function of the number of vertices.}
\label{ham}
\end{figure}

The Hamilton cycle problem has already been described. We randomly
generated one thousand graphs with the edge-to-vertex ratio such that 
$\approx 50\%$ of the graphs contained Hamilton cycles (crossover region).
The number of vertices ranged from 30 to 80. We used encodings of the 
problem as a $\datac$ program, logic program (in {\tt smodels} syntax) 
and as a propositional theory. {\tt dcs} performed better
than {\tt smodels} and {\tt smodels} performed significantly better
than {\tt csat} (Fig. \ref{ham}). We believe that 
a major factor behind poorer performance of {\tt csat} is that all known propositional
encodings of the hamilton cycle problem are much larger than those
possible with $\datac$ or logic programs (under the stable model
semantics). Propositional encodings, due to their size, rendered
{\tt csat} not practical to execute for graphs with more than 40
vertices.

\begin{figure}
\centerline{\hbox{\psfig{figure=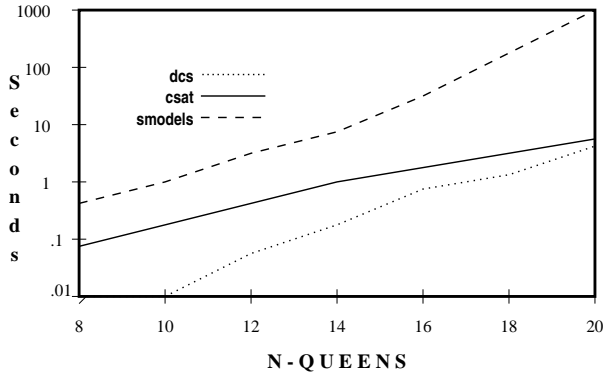}}}
\caption{$N$-queens problem; log scale}
\label{queen}
\end{figure}

The $N$-queens problem consists of finding a placement of $N$ queens on
an $N\times N$ board such that no queen can remove another. Both {\tt
csat} and {\tt dcs} execute in much less time than {\tt smodels} for these 
problems (Fig. \ref{queen}). Again the size of the encoding seems to be 
a major factor. One thing to consider in this case is that the number of 
rules for {\tt smodels} is approximately five times that for $\datac$ and 
more than twice that of propositional encodings.

\begin{figure}
\begin{center}
\begin{tabular}{|r|r|r|r|}
\hline
\multicolumn{1}{|c|}{B-N} &
\multicolumn{1}{|c|}{csat} &
\multicolumn{1}{|c|}{dcs} &
\multicolumn{1}{|c|}{smodel} \\
\hline
b-n&sec&sec&sec\\
\hline
3-13 &  0.03 & 0.00 & 0.12\\
3-14 &  0.05 & 0.00 & 0.16 \\
4-14 &  0.05 & 0.01 &  0.23 \\
4-43 &  0.59 & 1.91 & 5.23 \\
4-44 &  1.95 & 51.04 & 5.55 \\
4-45 &  1599.92 & 226.44  & 12501.00 \\
\hline \end{tabular} \end{center}

\caption{Schur problem; times and the number of choice points.} 
\label{schur}
\end{figure}

The Schur problem consists of placing $N$ numbers $1,2,\ldots,N$ in $B$ 
bins such that no bin is closed under sums. That is, for all numbers 
$x$, $y$, $z$, $1\leq x,y,z\leq N$, if $x$ and $y$ are the same bin, 
then $z$ is not ($x$ and $y$ need not be distinct). The Schur number
$S(B)$ is the maximum number $N$ for which such a placement is still
possible.
It is known to exist for every $B\geq 1$. We considered the problem of
the existence of the placement for $B=3$ and $N=13$ and $14$, and for
$B=4$ and $N=43, 44$ and $45$. In each case we used all three
systems to process the corresponding encodings. The results are shown in
Fig. \ref{schur}. It follows that $S(3)=13$ and $S(4)=44$. Again, {\tt
dcs} outperforms both {\tt smodels} and {\tt csat}.

\begin{figure}
\centerline{\hbox{\psfig{figure=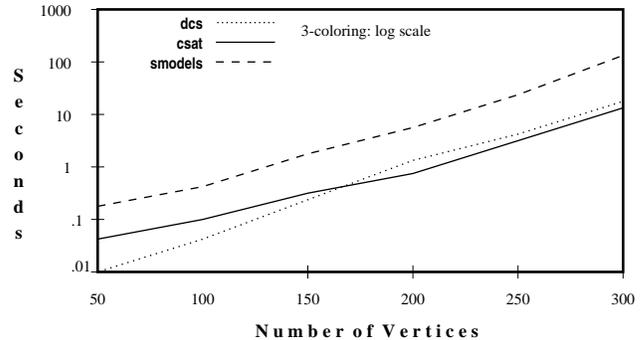}}}
\caption{3-coloring problem; log scale.}
\label{color}
\end{figure}

Results for graph 3-coloring for graphs with the number of vertices
ranging from $50$ to $300$ are shown in Fig. \ref{color} (for every
choice of the number of vertices, 100 graphs from the crossover region
were randomly generated). Both {\tt dcs} and {\tt csat} performed better
than {\tt smodels}.  Again the size of the 
theory seems to be a factor. The CNF theory for coloring is smaller 
than a logic program encoding the same problem. The sizes of propositional 
and $\datac$ encodings are similar.

Results for the pigeonhole placement problem show a similar performance
of all three algorithms, with {\tt csat} doing slightly better than 
the others and {\tt dcs} outperforming (again only slightly) {\tt
smodels}.  

\begin{figure}
\centerline{\hbox{\psfig{figure=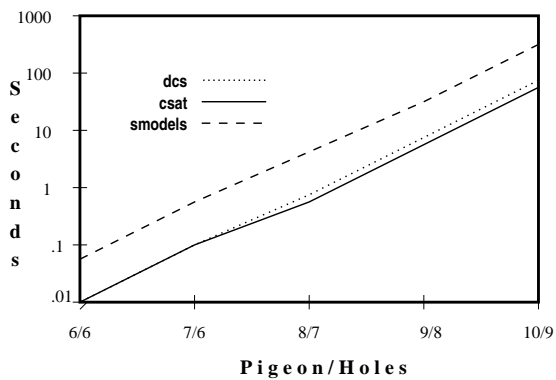}}}
\caption{Pigeonhole problem; log scale.}
\label{ph}
\end{figure}

\section{Conclusions}

We described a new system, $\datac$, for solving search problems. We
designed $\datac$ so that its semantics was as close as possible to that
of propositional logic. Our goal was to design a system that would
result is short problem encodings. Thus, we provided constructs for some
frequently occurring types of constraints and we built into $\datac$ 
elements of nonmonotonicity by including Horn rules in the syntax. As 
a result, $\datac$ programs encoding search problems are often much 
smaller than those possible with propositional theories. Experimental
results show that {\tt dcs} often outperforms systems based on 
propositional satisfiability as well as systems based on
nonmonotonic logics, and that it constitutes a viable approach to
solving problems in AI, constraint satisfaction and combinatorial
optimization. We believe that our focus on short programs is the
key to the success of $\datac$ and its reasoning engine {\tt dcs}.
Our results show that when building general purpose solvers of search
problems, the size of encodings should be a key design factor.

\end{document}